\documentclass[conference,a4paper]{IEEEtran}
\IEEEoverridecommandlockouts 
\ifCLASSINFOpdf
\else
\fi
\usepackage{amsmath}
\usepackage{tabu}
\usepackage{caption}
\usepackage{algorithm}
\usepackage{algorithmic}
\usepackage{graphicx}
\usepackage{tabu}
\begin{document}
%
\title{Generating Graphical Chain by Mutual Matching of Bayesian Network and  Extracted Rules of Bayesian Network  Using Genetic Algorithm }

\author{\IEEEauthorblockN{Mostafa Sepahvand$^1$\,
Ghasem Alikhajeh$^2$,
Meysam Ghaffari$^3$, 
Abdolreza Mirzaei$^4$ 
}
\IEEEauthorblockA{School of Electrical and Computer Engineering,
Isfahan University of Technology,
Isfahan, Iran\\ Emails: \tt $^1$ m.sepahvand@ec.iut.ac.ir, $^2$ g.alikhajeh@ec.iut.ac.ir}
\IEEEauthorblockA{ {\tt $^3$ meysam.ghafari@ec.iut.ac.ir, $^4$ mirzaei@cc.iut.ac.ir}}
}



\maketitle

\begin{abstract}
With the technology development, the need of analyze and extraction of useful information is increasing. Bayesian networks contain knowledge from data and experts that could be used for decision making processes But they are not easily understandable thus the rule extraction methods have been used but they have high computation costs. To overcome this problem we extract rules from Bayesian network using genetic algorithm. Then we generate the graphical chain by mutually matching the extracted rules and Bayesian network. This graphical chain could shows the sequence of events that lead to the target which could help the decision making process. The experimental results on small networks show that the proposed method has comparable results with brute force method which has a significantly higher computation cost.

\end{abstract}



%
\IEEEpeerreviewmaketitle

\section{Introduction}
Regarding to the information society which is relied on the extracted knowledge from intelligent systems and experts attract researchers in the field of  machine learning and knowledge discovery. This field helps us to deal with enormous data from variety of sources. This data contains hidden knowledge which could be valuable and helps us to improve decision making processes \cite{chan2012bayesian}. For example, existing data about sales of a company could contains useful relation between customers and products. Discovering these kind of relations could increase the sales and benefits of the companies. The growth of stored data is much more than the ability of human data analysis. 
Thus using Artificial Intelligence(AI) and machine learning techniques for extracting knowledge is inevitable.

Machine learning and AI methods try to discovering rules or a function from data. For instance SVM, Neural network and decision tree based methods that has been used widely in this scope. Combining Expert knowledge and the relation between attributes could improve the efficiency of the algorithm. Bayesian networks is one of the most known algorithms which has this capability \cite{adriana2013fuzzy,phukoetphim2013knowledge}.

Bayesian network (BN) is a learning method which combines the expert knowledge and training data \cite{pearl2011bayesian}. BN is a directed acyclic graph that its edges contains relation between nodes and nodes contains conditional probabilities of the node with its edges. Knowledge represented by BN and conditional probabilities are not understandable for the expert because it contains the complete information about nodes and the relation between them which make the network complex which most of them are not so effective.

 Extracting rules has been studied to overcome this problem. These extracted rules could be used for the classification and other applications either but they are suffering from not considering the chain relation between attributes. Thus we are going to generate graphical chains by matching extracted rules and BN. The graphical chain shows the path from each node to the target node by considering the relevant probabilities and choosing the most probable path. Knowing this most probable path we could specify the most effective node to the target node in each state which could be useful in decision making processes. Extracting rules from BN has a huge computational cost (That will be described in section III) which could not easily implemented Thus we use genetic algorithm to overcome this problem.
 
 The rest of this paper organized as follows. Section II is the related works, section III is the background of the BN. In the section IV we explain the proposed method. Section V is an example (Asia network) and section VI is the experimental results. Finally the conclusion is in section VII.

\section{Related Works}
Rule extraction methods usually try to extract a minimal network of rules instead of number of connected hidden units. Tickle et al. Describe detailed competed this subject and
provide a taxonomy \cite{tickle1998truth}. There are a variety of methods for extracting rules. Most of rule extraction methods use classifiers to extract rules. Methods based on Support Vector machines \cite{zhu2013rule} and Artificial Neural Networks \cite{chorowski2011extracting} are well-known methods in this field. Barakat and Diederich propose a method based on SVM which uses the trained SVM's support vectors to impel the decision tree \cite{barakat2005eclectic} which is based on eclectic approach. Method named OSRE for each training sample finds effective inputs and forms conjunctive rules by them \cite{etchells2006orthogonal} that is pedagogical approach. In this approach the internal structure of the classifier is irrelevant and just the input and output’s relation is evaluated.  Another type of methods are called decompositional methods that extract rules from structure of the network. For example Krishnan et al. use an optimizing minimizing search space technique using sorted weights \cite{krishnan1999search} and Re-RX method trains, analyze and prunes a neural network recursively and generates rules set hierarchically \cite{setiono2008recursive}. 

BNs having the knowledge from data and expert simultaneously and contain the relation between variables, but they are not easily interpretable by the expert.Thus they could be the base of interesting approaches in the Rule extraction methods. BN is used for modeling biomass-based weed-crop problem which The trained BN contain 4 nodes (variables) and 5 edges that each node has three states high, medium and low. then a set of 27 rules with most probability from the BN are extracted. The algorithm used for the rule extraction considered all nodes of BN with all its states \cite{bressan2007biomass}.

A model based on rule extraction from the BN and Naive Bayes proposed for biomass-based weed-crop. They used a pruning strategy to optimize each rule in the sequence of the class probability estimate \cite{bressan2007probability}. Rlue extraction from BN and Naive Bayes has been used for modeling the weed infestation risk \cite{bressan2009using}.

The extracted rule from a BN is used to construct the arguments which each of the antecedent and subsequent rules is only one variable. Their aim is to determine the pair of variables that are most probable to affect each other. They add an undercatter phase to break the rules which observes a variable between them that reduce the probability of the rule. This phase executes for each rule \cite{timmer2013inference}.
 
But the biggest problem of these techniques is the computational costs which grows exponentially \cite{tickle1998truth} and even some of them are NP complete problems which cause most solution applicable to small networks. To overcome this problem Genetic Algorithm methods has widely used because it is a heuristic method and it’s computational cost does not rely on the problem too much. Santos et al. apply genetic algorithm on  \cite{kohavi1997wrappers} and they use the rules quality as fitness function \cite{santos2000extracting}. Keedwell et al. go further and extract rules directly by genetic algorithm. They represent the input of neural network to the chromosomes which one means important and zero means don’t care \cite{keedwell2000creating}. This approach prospered by reducing the search space by Mohamed \cite{mohamed2011rules}. Besides that none of the mentioned methods extract chain rules that could be helpful in decision making processes especially when each attribute affects the other ones impact.

\section{Background}
\begin{figure}
\centering
\includegraphics[width=1.05\linewidth]{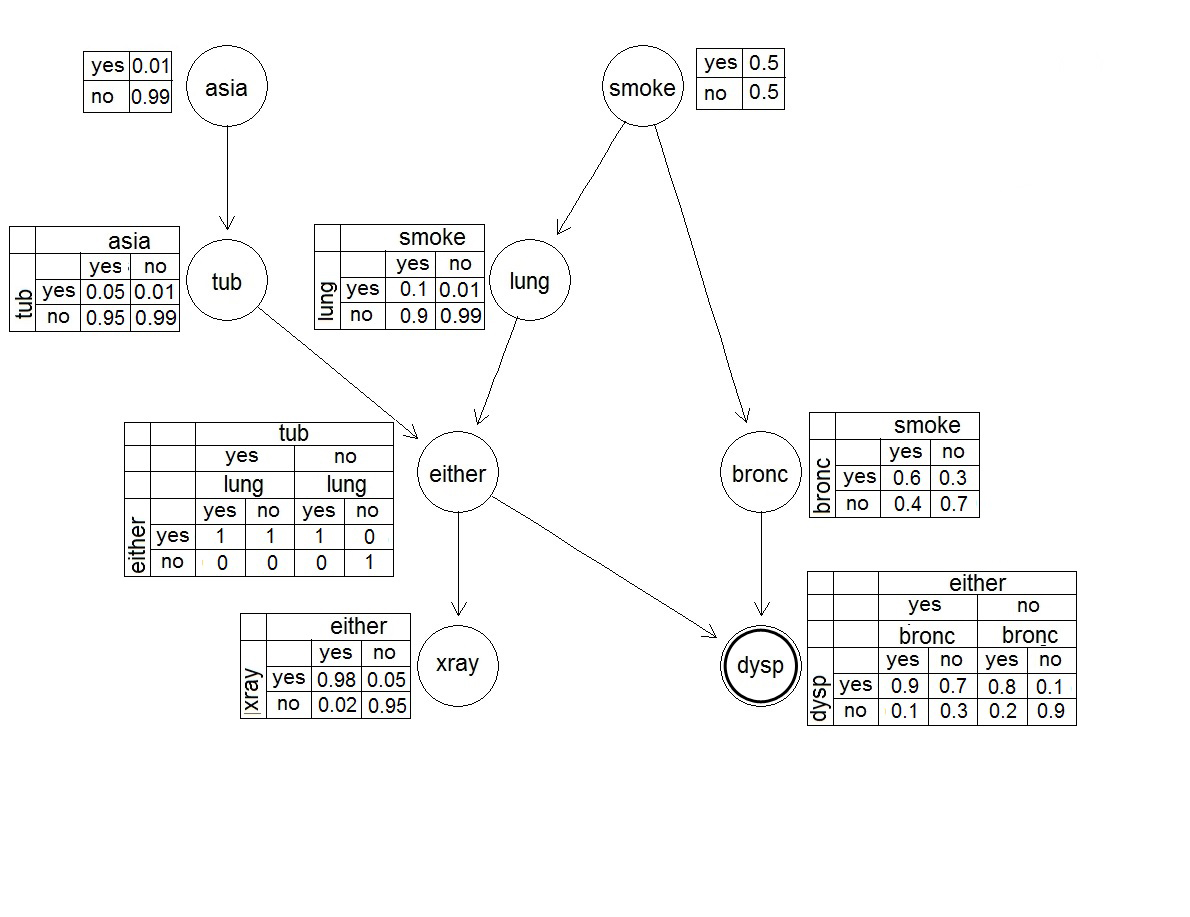}
\caption{Example of Bayesian network}
\label{fig:asiaP}
\end{figure}

the BN is a directed acyclic graph that nodes are the conditional probability of parents. The BN \textit{B} could be defined by $(G,\{{P_{X_1},…,P_{X_n}}\})$ that $G=(X,E)$ is an acyclic directed graph. Each $X_i$ is related to one probabilistic variable and ${{P_{X_1},…,P_{X_n}}}$ is the conditional probabilistic distribution (CPD) of the nodes of the graph. CPDs could be define as:
$P_{X_i}(X_i|Pa_g(X_i))$
That $Pa_g(X_i)$ specifies the parents of the $X_i$. In a BN the joint probability distribution is defined as equation \ref{equ:jointP} \cite{koller2009probabilistic}.

\begin{equation}
P_B(X_1,...,X_n)=\Pi_{i=1}^n P_{X_i} (X_i|Pa_g(X_i))
\label{equ:jointP}
\end{equation}
An illustrative example of a BN is depicted in Figure \ref{fig:asiaP}.

\section{Proposed Method}
The BN contain useful information and relationship between nodes But it suffers from complexity which comes from the complete relations between all nodes. BN contain lots of information, but most of them could not be used because in BN all events with different probabilities are depicted, but most of the time we are looking for most probable events. 

In these networks find the chains with higher probabilities which could be useful for experts or classification. In this paper we are going to extract effective chain rules which could be helpful to understand the events and predict the future events. For example in the medical field this knowledge could help the doctor to predict the future events which caused by present symptoms and by knowing the rules, prevent the deadly disease in the meantime.

Finding the optimum chain rules in BNs is a NP-complete problem, thus we use genetic algorithm to find these chain rules. To achieve this goal we first extract rules from BN with genetic algorithm and then extract the chain rules by matching extracted rules and BN.

\subsection{Genetic Algorithm}
The proposed genetic algorithm is based on \cite{ang2010evolutionary}. This algorithm has two part with different mutations and crossovers, Structural and local. Structural crossover will be applied and just the rules will be exchanged between chromosomes. Structural mutation could add or remove some rules from each chromosome either. In step two, local crossover and mutation will be done and the rules in the chromosome will be changed. 

The genetic algorithm starts with initialing the population. Then the fitness of chromosomes calculated and the parent chromosomes will be selected based on tournament selection method. Children will be generated by crossover and mutation on parent chromosomes. The survivor population will be selected from the best of children and parents. Finally a part of the best of the population will be sent to the local search procedure for improvement and the improved population will be sent to the next generation. In this part, we are going to briefly introduce the steps of the used genetic algorithm.The Pseudo-code of the genetic algorithm is depicted in the Algorithm 1.
\begin{algorithm}
\caption{Genetic  Algorithm For Rule Extraction}
\begin{algorithmic}
\STATE $\textbf{INPUT: } \alpha, genMax, Bayesian $\space$ Network $
\STATE $\textbf{OUTPUT: } Best $\space$ Chromosome$
\STATE $gen \leftarrow 1$
\STATE $Initial $\space $ Population $
\STATE $Calculate$\space $ the $\space $Fitness $\space $of $\space $the$\space $ Population   $
\WHILE{$gen $\space $ lower $\space $ than $\space $ genMax $}
\STATE $ \textbf{a.}	$\space$ select $\space $ pattern $\space $ p_1 $\space $ and $\space $ p_2 $\space $ from $\space $ population $\space $ using $\space $ tournament$\space $  selection $
\STATE $\textbf{b.}	$\space$apply $\space $ crossover $\space $ two $\space $ parents $\space $ and $\space $ generate $\space $ children's $\space $ c_1,c_2  $
\STATE $\textbf{c.}	$\space$mutation $\space $ for $\space $ each $\space $ c_1 $\space $ and $\space $ c_2 $
\STATE $\textbf{d.}	$\space$calculate $\space$ Fitness $\space$ c_1 $\space$ and $\space$ c_2$
\STATE $\textbf{e.}	$\space$add $\space $ c_1 $\space $ and $\space $ c_2 $\space $ to $\space $ population $
\STATE $\textbf{f.}	$\space$select $\space $ \alpha $\space $ percent $\space $ of $\space $ the $\space $ best $\space $ population $\space $ and $\space $ apply $\space $ local $\space $ search $\space $ on $\space $ them $ 
\STATE $\textbf{g.}	$\space$select $\space $ popsize $\space $ from population $\space $ and $\space $ send $\space $ to $\space $ next $\space $ generation$
\STATE $\textbf{h.} $\space$ gen \leftarrow gen+1$
\ENDWHILE
\STATE  $Best $\space$ Chromosome \leftarrow the $\space $ best $\space $ chromosome $\space $ from $\space $ the $\space $ last $\space $ population $
\label {algoritm:GA}
\end{algorithmic}
\end{algorithm}

\begin{figure}
\centering
\includegraphics[width=0.9\linewidth]{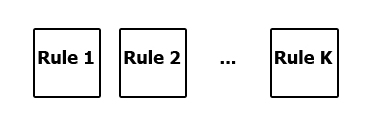}
\caption{Representation of chromosome}
\label{fig:chrom2}
\end{figure}

\begin{figure}
\centering
\includegraphics[width=0.9\linewidth]{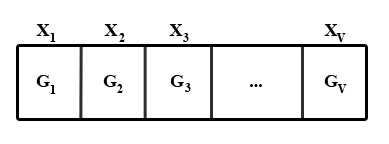}
\caption{Representation of Rule in chromosome}
\label{fig:gen2}
\end{figure}

\textbf{Representation:} Each chromosome contains a set of rules in the proposed method which means each gene in the chromosome represents a rule. The Chromosome structure is depicted in Figure \ref{fig:chrom2}. Number of rules in the chromosome is diverse and the genetic algorithm finds the best set. Each rule represent by a vector and it’s dimension is equal to the number of node as illustrated in Figure \ref{fig:gen2}. Each element could get its values based on the corresponding node. Rule representation shows in Figure \ref{fig:gen2} which $X_i$ is the node i and $G_i$ is its value. The value of node i could be define as equation \ref{equ:valGen}
\begin{equation}
G_i \in \{x_1,...,x_{s_i}, 0\}
\label{equ:valGen}
\end{equation}

that $\{$$x_1,..x_{s_i}$$\}$ is the possible states of $X_i$ node and $s_i$ is the number of states for node i. This means each node could get one of its states or zero that zero is neutral state which means the node has no effect in the rule. The target node in the BN could get one of its possible states and could not get zero. 

\textbf{Fitness Function:} The fitness of each chromosome is the sum of the fitness of all rules in that chromosome. If each rule is as $\phi \rightarrow \Psi $ then the fitness function could be calculated by equation \ref{equ:fitness}.

\begin{equation}
Fitness_{Rule set_j}=\Sigma_{R_i \in Ruleset_j} \sqrt{\frac{N-N_i}{N}} * fit_{R_i} * B(i)  
\label{equ:fitness}
\end{equation}
\begin{equation}
B(i)=\begin{cases}
1 $\space$  if $\space$ R_i \in $\{$ R_1,....,R_{i-1}$\}$ \\
0 $\space$ otherwise
\label{equ:Bi}
\end{cases} 
\end{equation}
That N is the number of rules, $N_i$ is number of rules with class label i and $Rule set_j$ is the rule set of chromosome j. The term $\sqrt{\frac{N-N_i}{N}}$ establishes the balance between numbers of rules in each class. B(i) is defined as equation \ref{equ:Bi} and used to eliminate the effect of duplicated rules in fitness calculation. The $fit_{R_i}$ is defined as equation \ref{equ:Ri}.
\begin{equation}
fit_{R_i}=(\frac{1}{(T_i+ \beta)^ \gamma}). P(\Psi | \Phi)
\label{equ:Ri}
\end{equation}

That $(\frac{1}{(T_i+ \beta)^ \gamma})$ inclines the algorithm to the general rules. The $T_i$ is the number of node in the antecedent of the $R_i$ that if we have more rules this term lower the fitness and thus we could achieve general rules. The $\beta$ is a number in [0,1] and $\gamma$ is the importance of general rules. By configuring $\beta$ and $\gamma$ we could tune the generality of the extracted rules and $P(\Psi | \Phi)$ is the probability of the target with evidence of the antecedent nodes that calculated from BN and $\Phi$ and $\Psi$ defined by equation 5.

\begin{equation}
\Phi \subset \{ X_1,...,X_V\}  $$\\$$
X_k \in \{ x_1,....,x_{s_k}\}  $$\\$$
\Psi \in \{ x_1,...,x_t\}
\end{equation}

That V is the number of nodes, $s_k$ is the number of states for $node_k$ and t is the number of states for the target node.

\textbf{Generating Initial Population:} The initial population will be generated randomly or based on the BN. In the random method the value of each node will be chosen randomly with uniform distribution. In the second method each value of the attribute will be chosen with the relevant probability in the network. Because we need the highly probable rules, using second method could lead us to the goal in lower time so we use BN in the initialization. 

In the assigning value method based on BN, each rule will be generated based on its probability which means more probable rules have more production chance and vise versa. The initialization is based on best first search. This procedure starts with the basic nodes (parent nodes) and specifies it’s state randomly determines with its child’s states. Then put its child in the queue and go to the next node. In this procedure, states with higher probability appear in more rules. Low probable values has the chance of existence either. For example in Figure \ref{fig:asiaP} the \textit{no} state for Asia node will be chosen for rules with chance of 0.99 and the \textit{yes} state will be chosen with the probability of 0.01. Assume that the Asia node gets the \textit{no} state, thus for the tub node the \textit{yes} has chance of 0.01 and the \textit{no} has chance of 0.99.

\textbf{Parent Selection:} Parents will be chosen by tournament selection method.  This method is based on suitably and each chromosome has the chance of selection based on its fitness.

\textbf{Crossover:} In this step first the parents rule set will be combined and then children will be generated randomly. Combined rule sets will be divided balanced between children. This means that generated children has almost equal rules.

\textbf{Mutation:} Mutation is defined by inserting and deleting rules. Deletion and insertion has the same probability and operator will choose randomly. Number of rules for insertion or deletion is based on normal distribution with zero average and the variance equal to one. Place of deletion will be choose randomly either. The value will be assigned to new rule based on the described method in the initialization.

\textbf{Local Search:} Part of best population will be sent to local search procedure for improvement. In the local search local mutation and crossover will be executed for the rules of each chromosome and the child will be generated. Parent pairs will be chosen  from chromosome rules randomly and children will be generated by one point crossover. Mutation is done by changing some of the rules of the chromosome. Rule index, Node index and number of changes will choose randomly. Finally if the fitness of the child is better than its parent, the parent will be replaced by the child. Local search procedure will execute \textit{genMaxLocal} times for all chromosomes. The procedure of local search is shown in Algorithm 2.

\begin{algorithm}
\caption{Local Search}
\begin{algorithmic}
\STATE $\textbf{INPUT: } population, GenMaxLocal $
\STATE $\textbf{OUTPUT: } changed $\space$ population$
\STATE $ Genlocal \leftarrow 1$ 
\WHILE{$Genlocal $\space $ lower  $\space $ than $\space $ is $\space $ not $\space $ met$}
\STATE $GenMaxLocal $
\FOR{$each $\space $ parent $\space $ in $\space $ The $\space $ population  $\space$  $}
\STATE $Local  $\space $ crossover  $\space $ on $\space $  parent  $\space $ and  $\space $ generate  $\space $ child $
\STATE $ Local $\space $ mutation $\space $ on $\space $ child$
\IF{$Fitness(child)>Fitness(parent)$}
\STATE $replace $\space $ child $\space $ with $\space $ parent$
\ENDIF
\ENDFOR
\STATE $ Genlocal \leftarrow Genlocal+1$
\ENDWHILE
\label {algoritm:LocalSearch}
\end{algorithmic}
\end{algorithm}

\subsection{Generating Graphical Chains}
In every extracted rule, non-zero values show nodes that have most effect on the target node. Each rule specifies that different nodes affect on the target node with which states. To extract the chain we use the union of all rule sets in the chromosome. For each effective node we add an edge with determined label from that node to its child in the path of that node to the target node (in the equivalent BN). Thus each rule specify some effective labeled edges and their union is a set of effective chain rules in the BN. The Pseudo code of creating graph that shows generating graphical chain rules is depicted in Algorithm 3.
\begin{algorithm}
\caption{Generating Graphical Chains}
\begin{algorithmic}
\STATE $\textbf{INPUT: } Rule$\space$ Set, Bayesian $\space$Network$
\STATE $\textbf{OUTPUT: } Graph$\space$ that$\space$ shows$\space$ effective $\space$ chains$
\STATE $Create $\space$ graph $\space$ with $\space$ size $\space$ nodes $\space$ of $\space$ Bayesian $\space$  Network $\space$ and $\space$ empty $\space$ edge $\space$ for $\space$ each $\space$ R_i $\space$ in $\space$ Rule$\space$ Set$
\FOR{$ each $\space$  R_i $\space$ in $\space$ Rule $\space$ Set $}
\FOR{$ each $\space$ e_i $\space$ (non $\space$ zeros $\space$ element $\space$ in $\space$ R_i $\space$ that $\space$ related $\space$ on $\space$ values $\space$ of$\space$  X_i) $}
\IF{$ not $\space$ observed $\space$ e_i $}
\STATE $ Add $\space$ edges $\space$ from$\space$  X_i $\space$ to $\space$ all $\space$ child $\space$ X_i $\space$ with $\space$ label $\space$  e_i $\space$ in $\space$ graph
			(define $\space$ child $\space$ X_i $\space$ using $\space$ Bayesian $\space$ Nework)$

\ENDIF

\ENDFOR

\ENDFOR
\end{algorithmic}
\end{algorithm}
\section{Example}
For better understanding of the proposed method we will describe the different steps of the proposed method on a sample network. The Asia BN has been chosen for this purpose. 
 Asia BN that has been generated from real medical data \cite{scutari2014package}. This network contains 8 nodes that described in Table \ref{tbl:DescAsia}.
 
 \begin{table} 
 \begin{center}
\centering
 \captionof{table}{Description of Asia nodes}
     \begin{tabular}{ |  p{3.5cm} | p{3.5cm} |}
     \hline
     Node(attribute) & Description \\
       \hline
      D (dyspnoea) & two-level factor (yes/no) \\
      \hline
      T (tuberculosis)  & two-level factor (yes/no)    \\
     \hline
      L (lung cancer)  & two-level factor (yes/no)    \\
     \hline
      B (bronchitis)  & two-level factor (yes/no)    \\
     \hline
      B (bronchitis)  & two-level factor (yes/no)    \\
     \hline
      A (visit to asia)	  & two-level factor (yes/no) \\
     \hline
      S (smoking)	  & two-level factor (yes/no) \\
     \hline
      X (chest X-ray)	  & two-level factor (yes/no) \\
     \hline
     E (tuberculosis versus lung cancer/bronchitis)	  & two-level factor (yes/no) \\
     \hline
     \end{tabular}
\label{tbl:DescAsia}
 \end{center}	
  \end{table}

We apply the genetic step of the proposed method on this network and extract the rules for the effective nodes on the dyspnoea(target node) with higher probabilities. some of these rules are illustrated in Table \ref{tbl:Rule}

\begin{figure}
\centering
\includegraphics[width=0.9\linewidth]{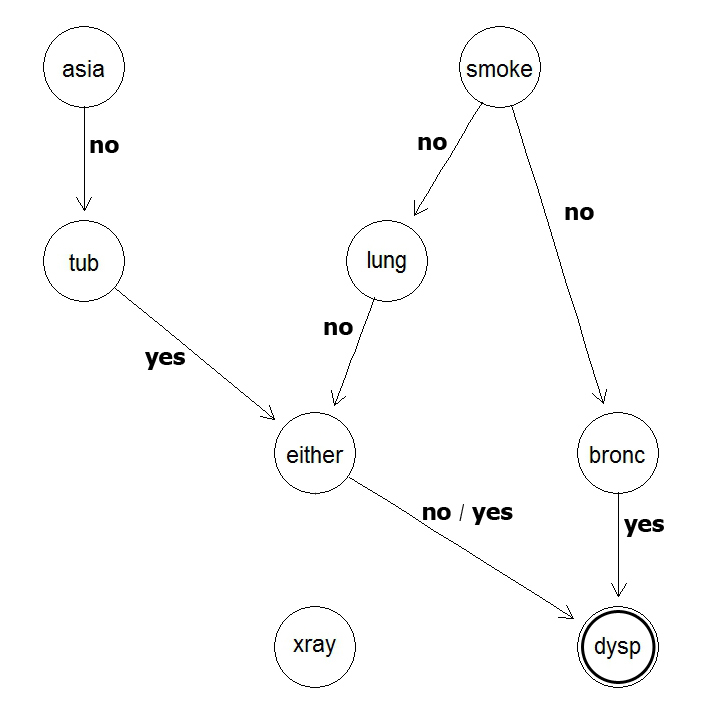}
\caption{The Extracted Graph}
\label{fig:asiaN5}
\end{figure}

 \begin{table}
 
 \begin{center}
 \captionof{table}{Chain Rules }
     \begin{tabular}{ |  p{5cm} | p{2.2cm} |}
     \hline
     Antecedent chain Rules (IF) & Result (THEN) \\
       \hline
       asia=no and either=yes & dysp=yes \\
      \hline
      bronc=yes and either=no  & dysp=yes    \\
     \hline
      bronc=yes  & dysp=yes    \\
     \hline
      either=yes  & dysp=yes    \\
     \hline
      asia=no and bronc=yes  & dysp=yes    \\
     \hline
     asia=no and tub=no and smoke=no and bronc=yes	  & dysp=yes \\
     \hline
     tub=yes and bronc=yes	  & dysp=yes \\
     \hline
     asia=no and smoke=no and bronc=yes and either=yes	  & dysp=yes \\
     \hline
    smoke=no and lung=no and either=yes  & dysp=yes \\
     \hline
      tub=no and smoke=yes and bronc=yes and either=no  & dysp=yes \\
          \hline
     \end{tabular}
\label{tbl:Rule}
 \end{center}
  \end{table}	

the result of extracting effective chains is illustrated in Figure \ref{fig:asiaN5}. As illustrated in Figure \ref{fig:asiaN5} the Path(Asia(no), tub(yes), Either(no), Then dyspnoea(yes)) shows that not being Asian and having tuberculosis and not having neither tuberculosis nor lung cancer with high probability will cause dyspnoea. 

\section{Evaluation}
there is not sufficient knowledge about appropriate rules, we use Asia, Cancer, Earthquacke, Sachs and Survey networks \cite{scutari2014package}. these networks are small enough that we could analyze them with brute force method. Table \ref{tbl:NetworkDetail}
\begin{table}
\begin{center}
 \captionof{table}{Details of Networks }
    \begin{tabular}{ | p{1.7cm} |p{1.7cm} | p{1.7cm} | p{1.7cm} |}
    \hline
    Network & Number of Nodes & Number of Arcs & Number of Parameters \\ \hline
    Asia & 8 & 8 & 18 \\ \hline
    Cancer & 5 & 4 & 10 \\ \hline
    Earthquake & 5 & 4 & 10 \\
    \hline
    Sachs & 11 & 17 & 178 \\
     \hline
    Survey & 6 & 6 & 21 \\
     \hline
    \end{tabular}
\label{tbl:NetworkDetail}
\end{center}
\end{table}

In the brute force method the rules will be generated with all possible values. on the other hands, the rules will be generated by all subsets of the nodes and applying this method is possible for small networks. To eliminate the not appropriate rules from the whole extracted rules, we will choose the rules that $P(\Psi | \Phi )$ is greater that specified threshold.
The following Table shows the average probability of the rules for the proposed method and the brute force method. For the brute force method, we use the threshold of 0.7. 
\begin{table}
\begin{center}
 \captionof{table}{Average Probabilistic of the rules for the proposed method and the brute force method }
    \begin{tabular}{ | p{1.7cm} |p{1.7cm} | p{1.7cm} | p{1.7cm} |}
    \hline
    Network & Proposed Method & Brute Force Method \\ \hline
    Asia & 0.8667 & 0.8564 \\ \hline
    Cancer & 0.9884 & 0.9620 \\ \hline
    Earthquake & 0.9537 & 0.9456 \\
    \hline
    Sachs & 0.8256 & 0.8133 \\
     \hline
    Survey & 0.5709 & 0.7 \\
     \hline
    \end{tabular}
\label{tbl:AvgProb}
\end{center}
\end{table}

As illustrated in the table IV, the proposed method works better than brute force for most of the cases (except Survey), which means we could extract the optimum rules. Note that for the brute force method we use threshold of 0.7 which cause this weakness. but the Survey has just one rule with 0.7 but our method find other rules thus the brute force has better results in this case.Also the proposed method could extract these results in a significantly lower time. 

\begin{figure}
\centering
\includegraphics[width=0.9\linewidth]{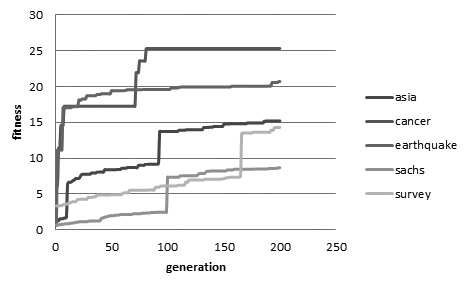}
\caption{Fitness Functions}
\label{fig:fitness}
\end{figure}

In the Figure \ref{fig:fitness} The fitness has been showed for the mentioned networks with 200 generation. Cancer and Earthquakes networks converge in lower than 100 iterations and Sachas and Asia converge in lower than 150 iterations. and Survey network converges in lower than 200 iterations.

\begin{figure}
\centering
\includegraphics[width=0.9\linewidth]{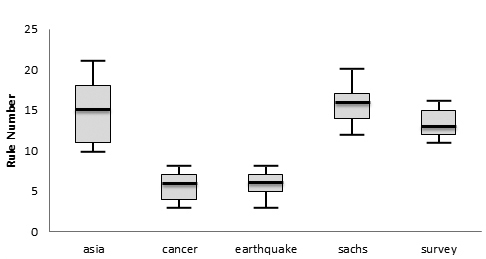}
\caption{Average number of rules}
\label{fig:tolombe}
\end{figure}

Figure \ref{fig:tolombe} shows the number of extracted rules by GA in 15 iterations. In this Figure, the median line in the box shows the average for multiple execution. Number of rules in the top quarter and below quarter of the median is depicted by the box. The maximum and minimum number of rules determined by horizontal lines. As it is illustrated, the dispersal of the proposed method for various network is low and it has similar results in multiple executions. 

\section{Conclusion}
In this paper we propose a novel method based on genetic algorithm for extracting graphical chain rules which could not be extracted from BN directly due to it's complexity. These rules could be useful for analyzing the effect of various factors on the disease and could help the experts to research on factors more focused. It also could help to predict the future events caused by current situation and symptoms. The evaluated results show that the proposed method works as well as brute force in small networks. Thus this proposed method could be used for big networks either because it do not suffer from the high complexity of computation of other methods. Thus we could expect that we could get reliable results on the big networks.





%

\bibliographystyle{plain}
\bibliography{bare_conf}

\end{document}